\documentclass[sigconf, language=english]{acmart}
\usepackage{multirow}
\usepackage{makecell}
\usepackage{bm}

\AtBeginDocument{%
  }

\copyrightyear{2025}
\acmYear{2025}
\setcopyright{cc}
\setcctype{by}
\acmConference[IVA Adjunct '25]{ACM International Conference on Intelligent Virtual Agents}{September 16--19, 2025}{Berlin, Germany}
\acmBooktitle{ACM International Conference on Intelligent Virtual Agents (IVA Adjunct '25), September 16--19, 2025, Berlin, Germany}\acmDOI{10.1145/3742886.3756728}
\acmISBN{979-8-4007-1996-7/2025/09}

\makeatletter
\ifcase\ACM@format@nr
\relax 
\or 
  \def\@authorfont{\large\sffamily}
  \def\@affiliationfont{\small\normalfont}
\or 
\or 
  \def\@authorfont{\LARGE\sffamily}
  \def\@affiliationfont{\large}
\or 
  \def\@authorfont{\LARGE}
  \def\@affiliationfont{\small}
\or 
  \def\@authorfont{\normalsize\normalfont}
  \def\@affiliationfont{\normalsize\normalfont}
\or 
  \def\@authorfont{\Large\normalfont}
  \def\@affiliationfont{\normalsize\normalfont}
\or 
  \def\@authorfont{\bfseries}
  \def\@affiliationfont{\mdseries}
\or 
  \def\@authorfont{\bfseries}
  \def\@affiliationfont{\mdseries}
\or 
  \def\@authorfont{\LARGE}
  \def\@affiliationfont{\large}
\or 
  \def\@authorfont{\large\sffamily}
  \def\@affiliationfont{\small\normalfont}
\fi
\makeatother

\begin{document}

\title[Towards Noise Reduction in SLP via Quaternions and Contrastive Learning]{Towards Skeletal and Signer Noise Reduction in Sign Language Production via Quaternion-Based Pose Encoding and Contrastive Learning}

\author{Guilhem Fauré}
\email{guilhem.faure@inria.fr}
\affiliation{
  \institution{Université de Lorraine, CNRS, Inria, LORIA}
  \city{F-54000 Nancy}
  \country{France}
}

\author{Mostafa Sadeghi}
\email{mostafa.sadeghi@inria.fr}
\affiliation{
  \institution{Université de Lorraine, CNRS, Inria, LORIA}
  \city{F-54000 Nancy}
  \country{France}
}

\author{Sam Bigeard}
\email{sam.bigeard@inria.fr}
\affiliation{
  \institution{Université de Lorraine, CNRS, Inria, LORIA}
  \city{F-54000 Nancy}
  \country{France}
}

\author{Slim Ouni}
\email{slim.ouni@loria.fr}
\affiliation{
  \institution{Université de Lorraine, CNRS, Inria, LORIA}
  \city{F-54000 Nancy}
  \country{France}
}

\begin{abstract}
One of the main challenges in neural sign language production (SLP) lies in the high intra-class variability of signs, arising from signer morphology and stylistic variety in the training data. To improve robustness to such variations, we propose two enhancements to the standard Progressive Transformers (PT) architecture (Saunders et al., 2020). First, we encode poses using bone rotations in quaternion space and train with a geodesic loss to improve the accuracy and clarity of angular joint movements. Second, we introduce a contrastive loss to structure decoder embeddings by semantic similarity, using either gloss overlap or SBERT-based sentence similarity, aiming to filter out anatomical and stylistic features that do not convey relevant semantic information.\\
On the \textsc{Phoenix14T} dataset, the contrastive loss alone yields a $16\%$ improvement in Probability of Correct Keypoint over the PT baseline. When combined with quaternion-based pose encoding, the model achieves a $6\%$ reduction in Mean Bone Angle Error. These results point to the benefit of incorporating skeletal structure modeling and semantically guided contrastive objectives on sign pose representations into the training of Transformer-based SLP models.
\enlargethispage{15pt}
\end{abstract}

\ccsdesc[500]{Computing methodologies~Natural language generation}

\keywords{Sign Language Production, Deep Learning, Contrastive Learning, Pose Encoding}

\maketitle

\section{Introduction}

Sign language is the main way of communication used in the deaf and hard-of-hearing (DHH) community. It leverages a wide range of manual (handshape, location, orientation, movement) and non-manual features (facial expression, body orientation, intensity) to convey ideas through a specific syntax and a rich vocabulary \cite{stokoe1976, stokoe1980, marcosreview2023}.\\
\indent With around 5\% of the global population affected by a disabling hearing loss, and a projection of over 700 million DHH people in 2050 according to the World Health Organization \cite{whodeafness2025}, it is essential to reduce the communication gap between deaf and hearing people, notably to avoid social exclusion. As a response to this growing need, a part of the research community has been working on developing new technologies for sign language recognition (SLR), sign language translation (SLT)---sign-to-text---and sign language production (SLP)---text-to-sign---tasks \cite{visicast2000, tessa2002, cuislr2017, camgozslt2020, stoll2018, saunders2020, fangsignllm2025}.\\
In the effort to develop digital tools that foster communication between deaf and hearing communities, the rise of deep learning has led to significant progress in recent years, particularly in SLR and SLT \cite{cuislr2017, puslr2018, camgozslt2018, camgozslt2020, lislt2020}. Since 2020, an increasing number of studies have focused on generating sign sequences from spoken language \cite{stoll2020, saunders2020, tang2024, baltazis2024, signdiff2025, signidd2025, he2025, walsh2024, signstokenszuo2024, yint2sgpt2024, mams2sl2024, fangsignllm2025, auralllm2025}. Among these, \citet{saunders2020} introduced the Progressive Transformers (PT) architecture, which has since emerged as a standard baseline in the field. 

\enlargethispage{15pt}

Despite these advances, SLP systems continue to face several fundamental challenges that hinder their usability in real-world applications. Key obstacles include the high intra-class variability of signs, the significant grammatical divergence between signed and spoken languages, and the scarcity of large-scale, annotated datasets with diverse vocabularies \cite{rastgooreview2021}. As a result, the generated outputs often lack the intelligibility, fluency, and naturalness required for deployment in practical communication scenarios.

In this work, we specifically address one of the core limitations of current SLP models: the visual variability of sign realizations, which introduces noise during training and impairs generalization. This variability arises primarily from two sources:
\begin{itemize}
    \item Inter-signers morphological differences, such as variations in bone lengths, which are not fully addressed by standard normalization techniques (e.g. (\citet{stoll2018}));
    \item Stylistic variations in the performance of a given sign or sentence---manifested through differences in amplitude, velocity, or positional noise---both across signers and within the same signer.
\end{itemize}

To mitigate the impact of these factors, we build upon the PT architecture and introduce two main contributions:

\begin{itemize}
    \item We represent skeletal poses using bone rotations encoded as quaternions rather than traditional 3D Cartesian joint coordinates, and replace the mean squared error (MSE) loss with a geodesic loss defined in quaternion space;
    \item We incorporate a contrastive loss into the training objective to structure the decoder's multi-head self-attention embeddings by pulling closer sequences with similar semantics and pushing apart dissimilar ones. We investigate two variants of this loss: one based on lexical overlap in the associated glosses (similar to the loss used in (\citet{walsh2024}) for the construction of their codebook), and another leveraging sentence Transformer embeddings (SBERT \cite{reimerssbert2019}) similarity scores between associated sentences, with the aim to capture subtler semantic relations.
\end{itemize}

These contributions aim to reduce the effect of non-semantic variability in training data and improve the semantic consistency and expressiveness of generated sign sequences.

\noindent We evaluate our approach on the widely used \textsc{Phoenix14T} dataset.\\
Code and demos are available online\footnote{\url{https://github.com/GFaure9/ContQuat-PT}}.

\section{Related Work}

\subsection{Sign Language Production}

Early approaches to SLP were primarily based on synthetic animation techniques relying on avatars and lookup tables containing pre-generated sequences for predefined sentences \cite{visicast2000, tessa2002, visicast2005, glauert2006vanessa, karpouzis2007educational}. These methods required the preparation and storage of a large set of sentence-sign pairs, making them costly and limiting their flexibility. Furthermore, the resulting avatar animations were often poorly received by the Deaf community due to their under-articulated, robotic, or unnatural movements \cite{uncannyvalley}.\\
\indent In recent years, progress in deep neural architectures has significantly advanced research in SLP. \citet{stoll2020} were the first to propose generating sign language pose videos from text using a three-stage pipeline: text-to-gloss conversion via a sequence-to-sequence model, motion graph-based sign stitching, and skeletal pose-to-video synthesis using a generative adversarial network (GAN). \citet{saunders2020} introduced a more streamlined autoregressive model that directly maps sentences or glosses to 3D skeletal poses using a Progressive Transformers architecture. They encode each frame’s temporal position in the sequence by appending a normalized counter value $\frac{t}{T}$ to its joint embedding.
Subsequent extensions of this model have incorporated data augmentation techniques (e.g., adding Gaussian noise, predicting multiple frames simultaneously), adversarial training, and mixture density networks \cite{saunders2021},  as well as skeletal graph self-attention mechanisms in the decoder \cite{skelgraph2021}. These improvements target the regression-to-the-mean effect in predicted signs and reduce error propagation during decoding. 
More recent approaches combine Transformer or diffusion-based architectures with vector quantization techniques to discretize the sign pose space \cite{walsh2024, signstokenszuo2024, yint2sgpt2024}. Typically, this involves a two-stage process: first, a Vector Quantized Variational Autoencoder (VQ-VAE) is trained to encode sequences of sign poses into discrete tokens by constructing a codebook; second, a model is trained to predict these tokens from textual input.\\
\indent While some recent models aim to generate avatar-based outputs \cite{baltazis2024, signstokenszuo2024}, the majority represent sign poses as 2D or 3D skeletal data and optimize a loss function based on the Cartesian coordinates of joints. However, this approach introduces several limitations:
\begin{itemize}
    \item The same sign performed by individuals with different body morphologies can lead to significant variations in joint coordinates;
    \item Computing the MSE over all joints tends to underweight the hands---critical for sign articulation---leading to reduced expressiveness;
    \item This representation ignores the underlying skeletal structure, requiring the model to implicitly learn limb dependencies, which may result in suboptimal performance.
\end{itemize}

To address these issues, some studies have introduced body-part-specific loss functions \cite{baltazis2024, signstokenszuo2024, tasyurek2025}, or additional representations using bone orientation vectors in $\mathbb{R}^3$, minimizing an MSE between predicted and reference orientations \cite{signidd2025}. However, the latter overlooks the non-Euclidian geometry of rotational space, and may misrepresent angular differences, limiting the precision needed to model fine-grained articulations.\\
\indent Finally, although pose tokenization effectively reduces stylistic variability \cite{walsh2024}, learning a robust codebook remains a non-trivial challenge, and may constrain the model’s ability to generate novel or unseen signs.

\subsection{Rotational Pose Encoding}

Instead of representing human motion as sequences of joint positions, an alternative is to describe it through bone rotations.  In this framework, each pose is reconstructed by recursively applying a sequence of bone rotations to a predefined skeletal structure in a resting ("T") pose, starting from the root joint. This representation helps prevent prediction errors caused by inconsistent bone lengths or anatomically implausible motions.\\
\indent Rotational pose encoding relative to a given skeletal structure has been used in various works on human motion recognition and prediction \cite{pavlovic2000, taylor2006, fragkiadaki2015, pavlloquaternet2018}. Rotations can be parameterized in several ways, including 3×3 rotation matrices, Euler angles, exponential maps, and quaternions \cite{grassiaexpmap1998, hanrotationreview2016}. However, many of these parameterizations present disadvantages for deep learning applications. For example, rotation matrices require enforcing six nonlinear constraints to remain within the 3D rotation group $SO(3)$, while Euler angles are prone to {\it gimbal lock} when two rotation axes become aligned, leading to the loss of one degree of freedom. More broadly, since $\mathbb{R}^3$ cannot be smoothly mapped to $SO(3)$, exponential maps may also lead to singularities.\\
\indent Unit quaternions offer a robust and efficient alternative by representing rotations in 4D space, avoiding these common pitfalls. They are numerically stable, support smooth interpolation, and simplify the composition of rotations  \cite{grassiaexpmap1998}.
Quaternions have been successfully employed in recurrent models for human motion understanding \cite{pavlloquaternet2018}, and more recently in sign language processing to construct sign language action embeddings \cite{wen2stagescore2024}.\\
Despite their strengths, quaternions---like all representations in four or fewer dimensions---are inherently discontinuous representations of 3D rotations, as shown in (\citet{zhoucontinuity2019}). The authors demonstrate that continuous representations of 3D rotations can be defined in 5D and 6D, making them better suited for learning. However, quaternions remain an attractive choice for our application, due to their compact 4D representation and low computational overhead while resolving common issues of classical 3D rotation representations.

\enlargethispage{10pt}

\subsection{Contrastive Learning}
\looseness-1  Contrastive learning is a machine learning paradigm in which models learn more effective representations by comparing samples—pulling positive pairs (similar samples) closer together in the embedding space, while pushing negative pairs (dissimilar samples) further apart \cite{lecun2006, khoslasupcont2020, selfsupcont2021}. This approach has proven successful in enhancing language embeddings \cite{gaosimcse2021}, visual representations \cite{chenvisualrepr2020}, and in aligning cross-modal embeddings \cite{openaiclip2021}. In the field of sign language technologies, contrastive learning has been primarily explored in sign-to-text translation frameworks \cite{ye2024improving, jiangsignclip2024, zhouslt2023, linslt2023}. \citet{ye2024improving} demonstrate that reducing the density of the sign pose representation space via contrastive learning improves SLT performance. In (\citet{jiangsignclip2024}) and (\citet{zhouslt2023}), contrastive learning is employed for visual-language pretraining by encouraging alignment between visual and textual embeddings when they correspond to matching $(\textit{ground truth}, \textit{label})$ pairs. \citet{linslt2023} supervise the learning of visual feature embeddings using Contrastive Concept Mining (CCM): a method that identifies "anchor words" from batch-level sentences and treats two sign sequences as a positive pair if both contain the same anchor word. A similar technique is adopted in (\citet{walsh2024}) for codebook training, where positive and negative pairs are constructed based on gloss overlap.\\
\indent Our approach draws inspiration from these works but applies contrastive losses directly within the latent space of the decoder's self-attention layers in the PT architecture. We hypothesize that aligning these latent representations to the underlying semantic distribution of the text before cross-modal attention encourages more efficient learning, by filtering out visual features that are not semantically relevant. This aligns with the motivation of (\citet{walsh2024}), which seeks to reduce signer-specific variability and promote person-invariant representations in sign language generation models.

\section{Methodology}

\subsection{Overview}

We adopt the PT architecture of Saunders et al. \cite{saunders2020} as our backbone, as it is a widely used baseline in SLP and offers a complete, publicly available implementation\footnote{\url{https://github.com/BenSaunders27/ProgressiveTransformersSLP}}.

\begin{figure}[h]
  \centering
  \includegraphics[trim=0cm 0.2cm 0.85cm 0cm, clip, width=\linewidth]{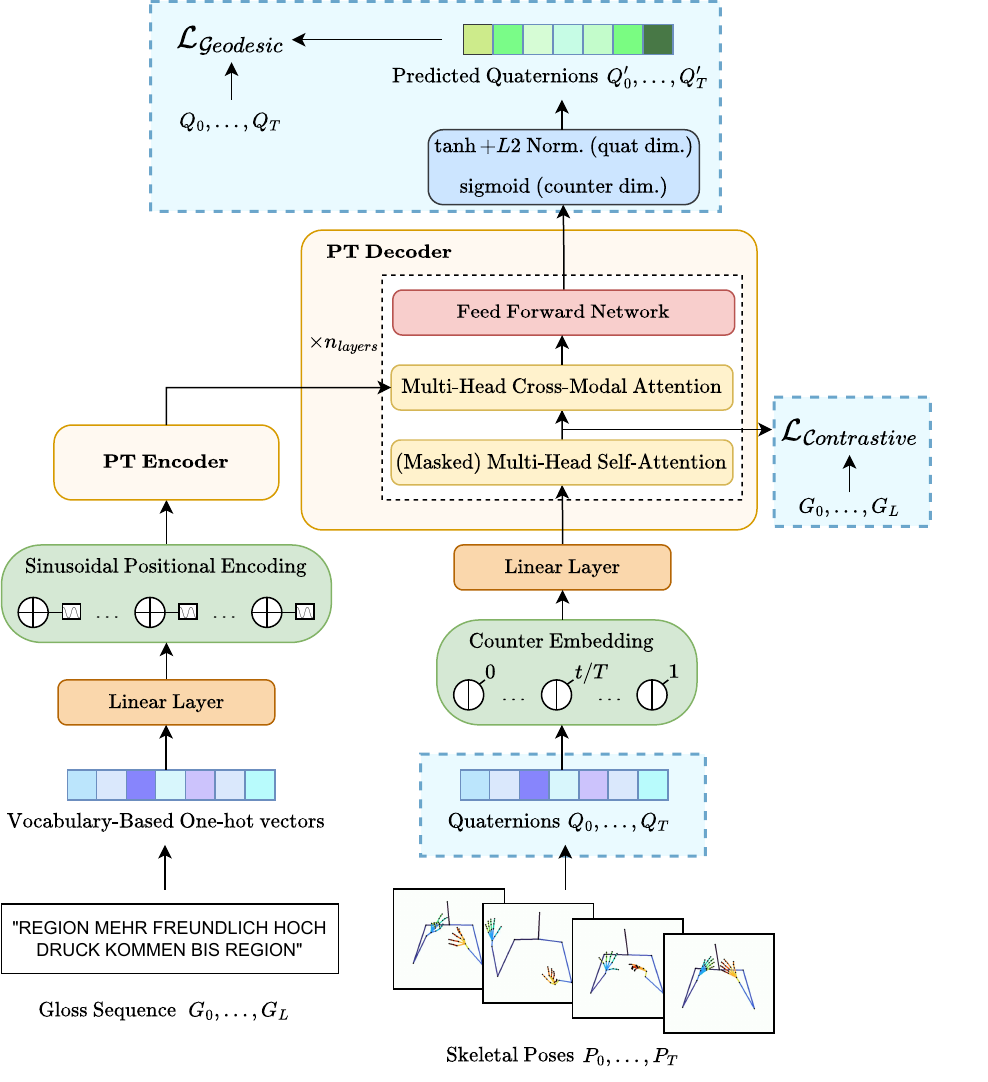}
  \caption{PT model architecture integrating quaternion-based pose encoding and supervised contrastive loss. Blue dotted boxes indicate the modules specific to our contributions.}
  \label{fig:overview}
\vspace*{-12pt}
\end{figure}

As shown in Figure \ref{fig:overview}, we propose two extensions:
(1) pose sequences are encoded via bone rotations using unit quaternions, replacing MSE loss on Cartesian coordinates with a loss based on the more natural geodesic norm;
(2) we explore two contrastive objectives, supervised by textual input, applied to the decoder's self-attention latent space to guide its structure to reflect semantic relationships. The first variant follows the loss formulation of \cite{walsh2024}, defining positive and negative pairs based on shared gloss presence. The second aligns similarity matrices between latent features and SBERT sentence embeddings.

\enlargethispage{10pt}

\subsection{Quaternion-Based Representation of Skeletal Poses}

From a skeletal pose in 3D Cartesian coordinates $P := (X_1, ..., X_{N_{\text{\scriptsize {joints}}}})$ lying in $\mathbb{R}^{N_{\text{\scriptsize {joints}}}\times3}$, given the graph structure of the skeleton and a reference T-pose $P_0$, we compute the 3D rotation in quaternions representation of the $i$-th bone as follows:

\begin{equation}
    q_i := \left( \cos(\theta_i/2), \sin(\theta_i/2) u^{(i)} \right) \in [-1, 1]^4
\end{equation}

$$
    \text{where }
    \begin{cases}
    \theta_i = \arccos(v^{(i)} \cdot v^{(i)}_0)\\
    v^{(i)} = \frac{X_{\text{Child}_i} - X_{\text{Parent}_i}}{\| X_{\text{Child}_i} - X_{\text{Parent}_i} \|}, ~  v^{(i)}_0 
    = \frac{X_{\text{Child}_i}^0 - X_{\text{Parent}_i}^0}{\| X_{\text{Child}_i}^0 - X_{\text{Parent}_i}^0 \|} \\
    u^{(i)} = \frac{ v^{(i)} \times v^{(i)}_0 }{\| v^{(i)} \times v^{(i)}_0 \|} \quad \text{('$\times$': classical cross product)}
    \end{cases}
$$

\begin{figure}[h]
  \centering
  \includegraphics[trim=0cm 0cm 0.7cm 0cm, clip, width=0.9\linewidth]{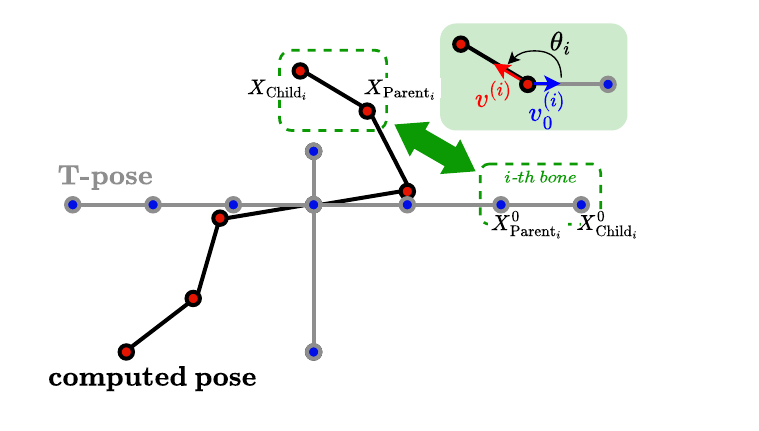}
  \vspace{-0.6cm}
  \caption{Illustration of bone rotation angle relative to a reference T-pose.}
  \label{fig:angle_t_pose}
\vspace*{-10pt}
\end{figure}

\noindent Hence, for each sequence of poses $\textbf{Y}:=(P_0, ..., P_T)$, we obtain the corresponding sequence of bones rotations $\textbf{R}:=(Q_0, ..., Q_T)$ where $Q_t := (q_1[t], ..., q_{N_{\text{\scriptsize bones}}}[t])$.\\
\indent Based on the definition of the geodesic distance between unit quaternions, we define the loss function between predicted rotations $\textbf{R}'$ and ground truth rotations $\textbf{R}$ as:

\begin{equation}
    \mathcal{L}_{\text{\scriptsize Geo}} := \frac{1}{(T+1)N_{\text{\scriptsize bones}}}\sum_{t=0}^T \sum_{i=1}^{N_{\text{\scriptsize bones}}} \arccos \left( 2 (q_i'[t] \cdot q_i[t])^2 - 1 \right)
\end{equation}

\noindent This loss temporally averages the mean rotation angle between predicted and ground truth bone orientations.\\
\indent Additionally, to enable reconstruction of predicted skeletal pose sequences in 3D Cartesian coordinates by recursively applying rotations from the root joint, we also predict the head node's position by minimizing the following MSE during training:

\begin{equation}
    \mathcal{L}_{\text{\scriptsize Root}} := \frac{1}{T+1} \sum_{t=0}^T \| X_{\text{\scriptsize Root}}'[t] - X_{\text{\scriptsize Root}}[t] \|_2^2
\end{equation}

\noindent In constrast, the PT baseline relies exclusively on an MSE loss over joint positions.

\subsection{Contrastive Losses}

The proposed contrastive losses are incorporated as regularization terms in the overall training objective, in addition to the standard SLP loss---either the MSE on joint positions or $\mathcal{L}_{\text{\scriptsize Geo}} + \mathcal{L}_{\text{\scriptsize Root}}$. A scaling factor $\lambda$ balances the contribution of the contrastive loss:
\begin{equation}
\mathcal{L}_{\text{\scriptsize Total}} := \mathcal{L}_{\text{\scriptsize SLP}} + \lambda \mathcal{L}_{\text{\scriptsize Cont}}
\end{equation}
The two contrastive losses we evaluate are presented in the following subsections.

\enlargethispage{10pt}

\subsubsection{Supervision with Glosses}

\begin{figure}[h]
  \centering
  \includegraphics[width=0.9\linewidth]{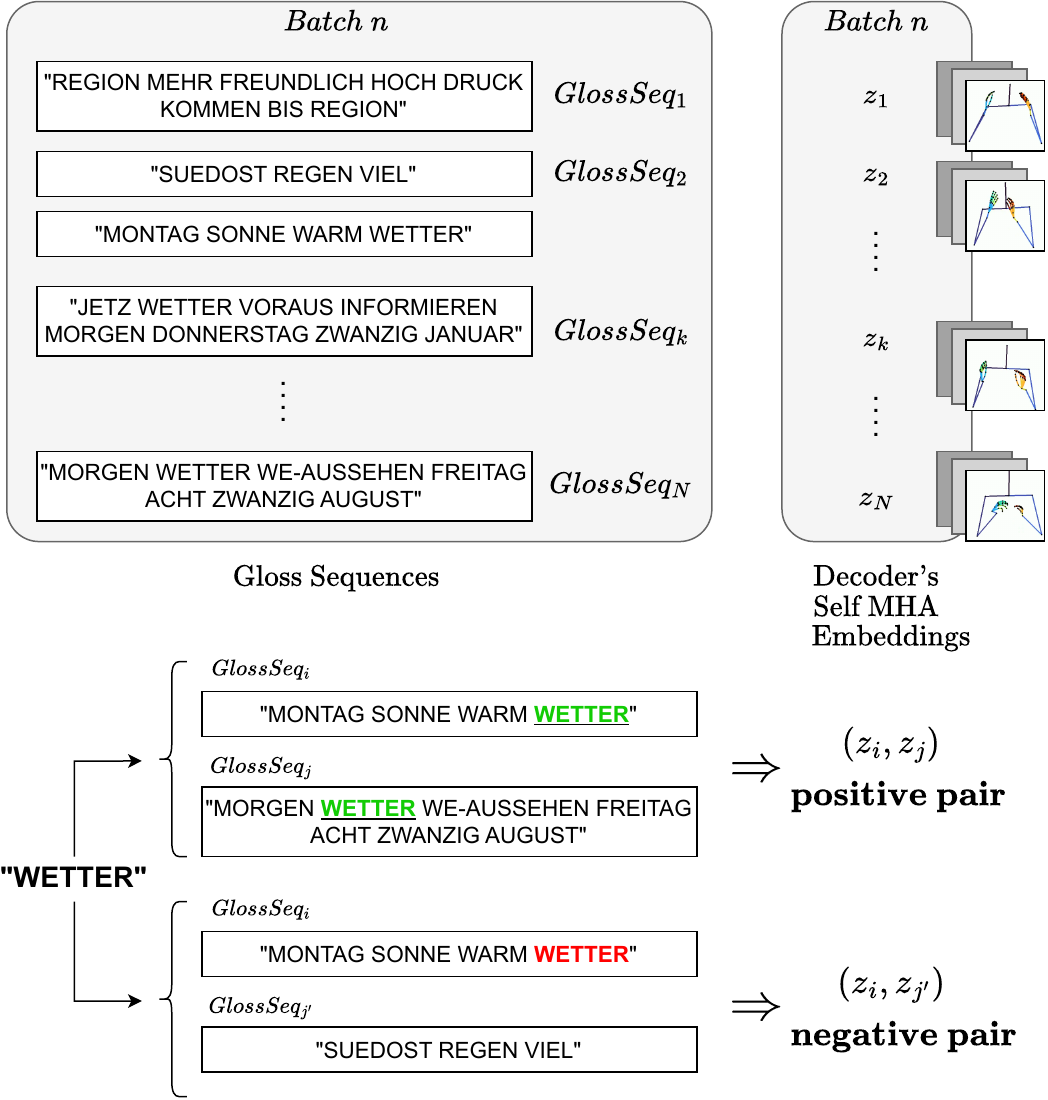}
  \caption{Definition of positive and negative pairs for the computation of $\mathcal{L}_{{\text{\scriptsize GlossSupCont}}}$.}
  \label{fig:contrastive_pairs}
\vspace*{-10pt}
\end{figure}

To define the supervised contrastive loss based on input gloss sequences, we follow the method proposed in (\citet{walsh2024}), itself inspired by (\citet{khoslasupcont2020}). For each batch of decoder's self-attention hidden representations, we first extract the set of all unique gloss tokens (referred to as anchors, indexed by $I$) appearing in the batch. For each anchor $i \in I$, we identify within the batch the sequence where $i$ occurs most frequently. This sequence will serve as the reference. The remaining sequences are split into \textit{positives} $A(i)$ (those that also contain $i$) and \textit{negatives} $B(i)$ (those that do not), as illustrated in Figure \ref{fig:contrastive_pairs}.

\noindent Based on this grouping, we define the following contrastive loss over the $l$-th self-attention layer’s output $\textbf{Z}_{\text{\scriptsize batch}}^l := (z_1, ..., z_N)$, where $N$ is the batch size:

\begin{equation}
    \mathcal{L}^{(l)}_{\text{\scriptsize GlossSupCont}} := -\sum_{i \in I}\log \left( \frac{1}{|A(i)|} \sum_{a \in A(i)} \frac{\exp(\frac{z_{f(i)} \cdot z_a}{\tau})}{ \sum_{b \in B(i)} \exp(\frac{z_{f(i)} \cdot z_b}{\tau})} \right)
\end{equation}

$$
    \text{with }
    \begin{cases}
    f(i) := \arg \max_{k}\{\sum_{m \in k\text{\scriptsize -th Gloss Sequence}} \textbf{1}_{m=i}\} \\
    A(i) := \{ a \in \llbracket 1, N \rrbracket ~ | ~ i \in a\text{\scriptsize -th Gloss Sequence} \} \backslash \{f(i)\} \\
    B(i) := \{ b \in \llbracket 1, N \rrbracket ~ | ~ i \notin b\text{\scriptsize -th Gloss Sequence} \}
    \end{cases}
$$

\noindent We average the per-layer losses to obtain the final objective:

\begin{equation}
    \mathcal{L}_{\text{\scriptsize GlossSupCont}} := \frac{1}{n_{\text{\scriptsize layers}}} \sum_{l=1}^{n_{\text{\scriptsize layers}}} \mathcal{L}^{(l)}_{{\text{\scriptsize GlossSupCont}}}
\end{equation}

\subsubsection{Supervision with SBERT Embeddings}

\begin{figure}[h]
  \centering
  \includegraphics[trim=0cm 0cm 0.8cm 0cm, clip, width=\linewidth]{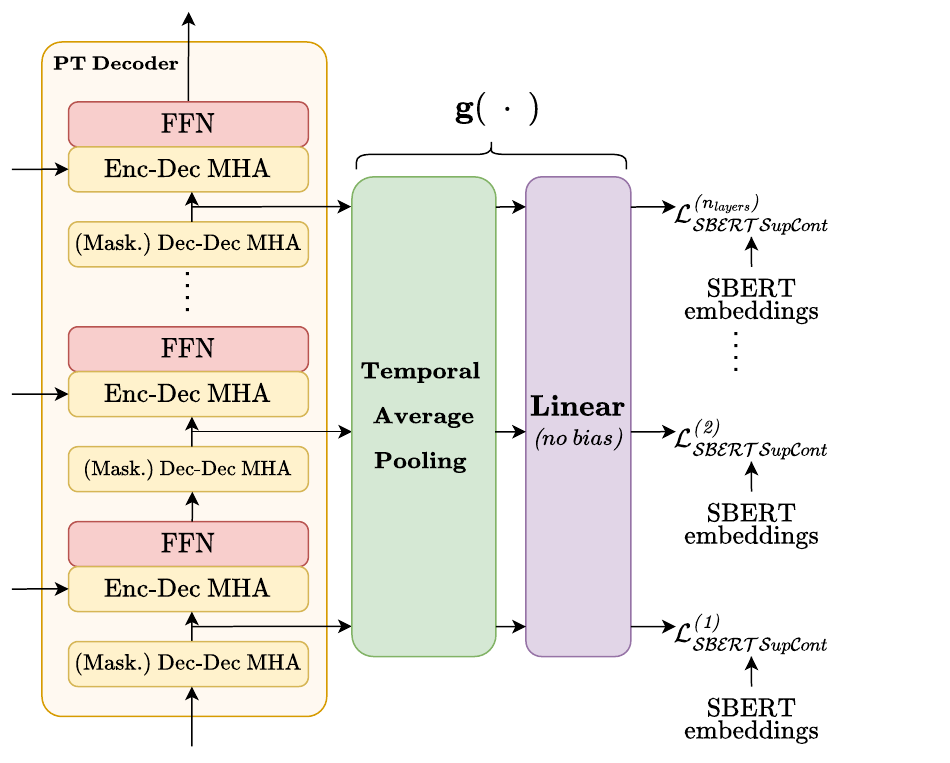}
  \vspace{-0.6cm}
  \caption{Projection into latent space prior to $\mathcal{L}_{\text{\scriptsize SBERTSupCont}}$ computation. Decoder's self-attention outputs are dimensionally aligned with SBERT embeddings.}
  \label{fig:dim_reduction}
\end{figure}

To incorporate finer knowledge of semantic relationships between sequences embeddings, we build an alternative loss based on the cosine similarity between input sentences once embedded via a sentence Transformer (SBERT) \cite{reimerssbert2019}. Upstream, the embeddings $(\text{SBERT}_k)_k$ of input sentences are thus computed using the 'all-MiniLM-L6-v2' model from the \textit{Hugging Face} library\footnote{\url{https://huggingface.co/sentence-transformers/all-MiniLM-L6-v2}}. These embeddings are of dimension 384. Hence, to match the SBERT embedding size before computing the loss, the outputs of the decoder’s self-attention blocks are first averaged along the temporal dimension via average pooling, and then projected through a linear layer (cf. Figure \ref{fig:dim_reduction}).

\noindent These previous steps enable the computation of the following loss over the batch output of the $l$-th self-attention layer:

\begin{equation}
    \mathcal{L}^{(l)}_{\text{\scriptsize SBERTSupCont}} := \frac{N(N-1)}{2} \sum_{1\leq i<j \leq N} d_{i, j}^2
\end{equation}

\begin{equation}
    \text{with } d_{i, j} :=  \textbf{sim}(\textbf{g}(z_i), \textbf{g}(z_j)) - \textbf{sim}(\text{SBERT}_i, \text{SBERT}_j)
\end{equation}

\begin{figure}[h]
  \centering
  \vspace{-0.3cm}
  \includegraphics[trim=0cm 0.5cm 0.5cm 0.8cm, clip, width=0.8\linewidth]{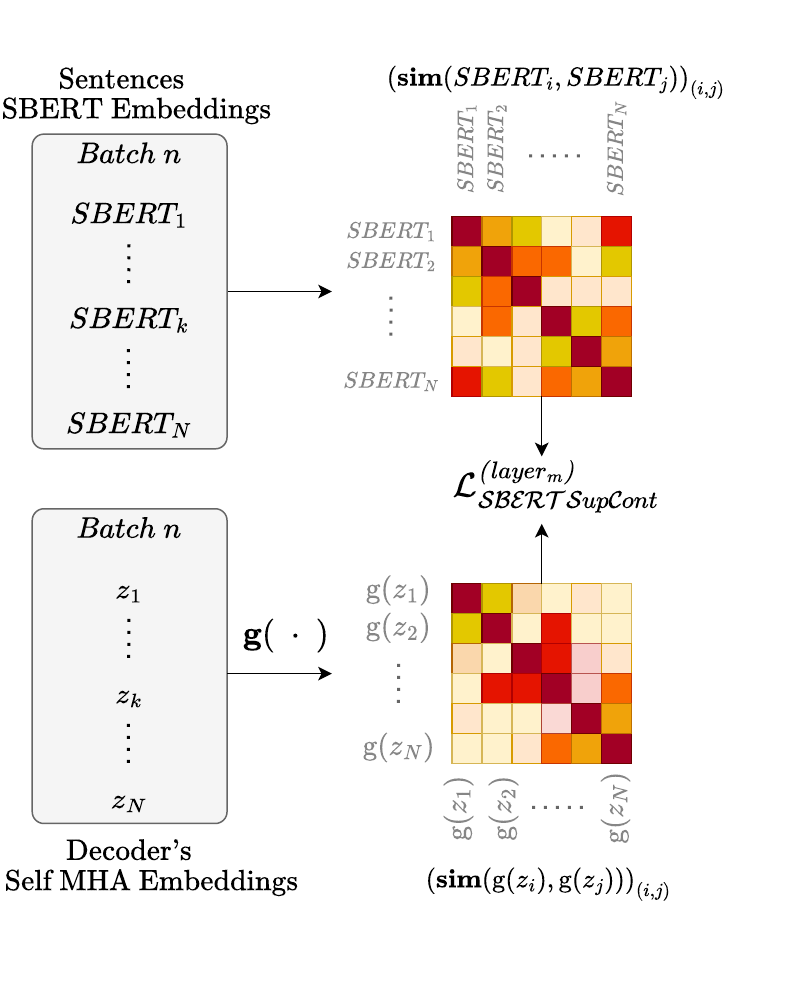}
  \vspace{-0.8cm}
  \caption{Computation of similarity matrices of SBERT and batch samples $\textbf{g}(z)$ for $\mathcal{L}_{\text{\scriptsize SBERTSupCont}}$.}
  \label{fig:sbert_loss}
  \vspace{-0.1cm}
\end{figure}

\noindent Where $\textbf{sim}(x, y) := \frac{x \cdot y}{\|x\| \|y\|}$ denotes cosine similarity, and $\textbf{g}(\cdot)$ is the transformation applied by the projection layers. The goal is to align the similarity matrices computed from pose embeddings and SBERT embeddings (see Figure \ref{fig:sbert_loss}), such that the resulting latent spaces are structured according to semantic relationships, while minimizing the influence of non-semantic features.

\noindent As for the first variant of contrastive objective, the overall loss is finally obtained by averaging accross all layers:

\begin{equation}
    \mathcal{L}_{\text{\scriptsize SBERTSupCont}} := \frac{1}{n_{\text{\scriptsize layers}}} \sum_{l=1}^{n_{ \text{\scriptsize layers}}} \mathcal{L}^{(l)}_{\text{\scriptsize SBERTSupCont}}
\end{equation}

\section{Experiments}
\subsection{Experimental Settings}

\subsubsection{Dataset}

We train and evaluate our models on the widely used \textsc{Phoenix14T} dataset, which comprises 8,257 sequences of German Sign Language (DGS) performed by 9 signers, covering 1,066 glosses and a vocabulary of 2,887 unique words \cite{rastgooreview2021}. While limited, this dataset is a standard benchmark in SLP, making it a reliable starting point for assessing model performance and comparison with existing methods before scaling to richer datasets.

\subsubsection{Preprocessing}

3D skeletal coordinates are extracted using \textsc{MediaPipe}’s pose and hand landmarks detection\footnote{https://github.com/google-ai-edge/mediapipe}. Joint positions are then refined following the method of Zelinka and Kanis \cite{zelinka2020}, which interpolates missing joints and applies inverse kinematics to correct misplacements while preserving bone length consistency. Finally, skeletons are normalized as in (\citet{stoll2018}), based on shoulder-to-shoulder distance to reduce size variation across subjects.\\
\indent We use gloss sequences as input for the generation process, following the original PT paper \cite{saunders2020} and subsequent works \cite{signidd2025, walsh2024, g2pddm2024}.

\subsubsection{Evaluation Metrics}

We evaluate the tested configurations using standard metrics in SLP to quantify the alignment between generated and reference skeletons. Specifically, we compute the Mean Joint Error (MJE), defined as the Euclidian distance between predicted and ground truth joints averaged over all joints and times steps, as used in prior work \cite{baltazis2024, walsh2024, signidd2025}. Similarly, we define the Mean Bone Angle Error (MBAE) as the mean angular deviation (in degrees) between predicted and reference bones. It quantifies articulation accuracy independently of bone length, which makes it particularly appropriate for evaluating our quaternion-based variant.\\
\indent We also compute the Probability of Correct Keypoint (PCK), which measures the proportion of predicted joints falling within a joint-specific neighborhood of their corresponding ground truth positions in the image plane. This neighborhood is defined for each joint, projected onto the $(x, y)$ plane, as a threshold $\alpha$ of the radius of its bounding disk. As in (\citet{kapoor2021}), we choose $\alpha=0.2$. This metric accounts for varying spatial scales across different body parts.\\
\indent For both MJE and PCK, sequences are first aligned using Dynamic Time Warping (DTW) based on Euclidean distance between joint coordinates. For MBAE, the applied DTW is instead computed to minimize angular differences between corresponding bones.\\
\indent It is worth noting that our focus is ultimately on the relative changes in the reported metrics with respect to the PT baseline, rather than on their absolute values. This approach allows us to assess whether the proposed changes lead to measurable improvements within PT-like architectures.

\subsubsection{Implementation Details}

We retain the original configuration of the PT model from the reference repository, using 2 layers, 4 attention heads, and an embedding size of 512 for both the encoder and decoder. The temperature parameter for the $\mathcal{L}_{\text{\scriptsize GlossSupCont}}$ is set to $\tau=1$. Moreover, for $\mathcal{L}_{\text{\scriptsize GlossSupCont}}$, we set the scaling factor to $\lambda=10^{-4}$ and the batch size to $64$.\\
\indent Training is conducted on an NVIDIA GeForce RTX 2080 Ti GPU. No significant computational overhead is observed between the baseline and the quaternion-based variant, with training times averaging $\sim$1h50 for 1000 epochs. Adding contrastive losses substantially increases runtime, requiring $\sim$3h30 with $\mathcal{L}{\text{\scriptsize SBERTSupCont}}$ and up to $\sim$11h with $\mathcal{L}{\text{\scriptsize GlossSupCont}}$.
\subsection{Results}

\subsubsection{3D Cartesian Positions VS Quaternion-based Rotations}

\begin{figure}[h]
  \centering
  \includegraphics[width=\linewidth]{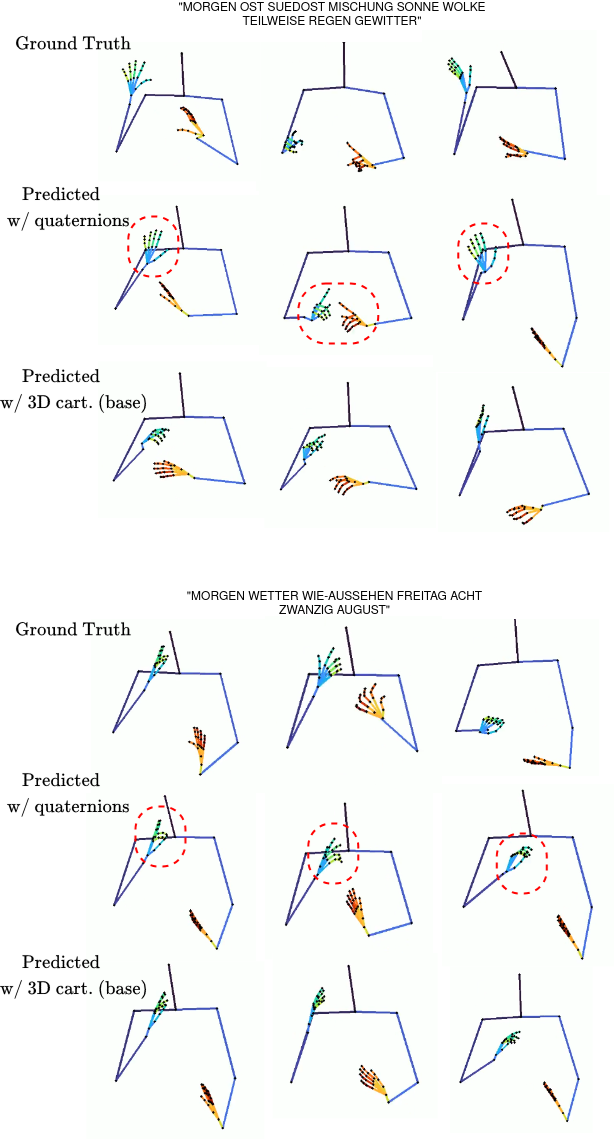}
  \caption{Qualitative comparison of ground truth and predicted skeletal poses with the base PT model and its quaternion-based variant.}
  \vspace{-0.5cm}
  \label{fig:qualitative_gt_vs_base_vs_quat}
\end{figure}

\begin{table}[t]
  \caption{Evaluation metrics on \textsc{Phoenix14T} test set for different configurations. Values are reported as $\textsc{Mean}^{\pm \textsc{std}}$. A \textbf{bold} score indicates the best result. The second-best result is \underline{underlined}.}
  \label{tab:results}
  \centering
  \begin{tabular}{lcccc}
    \toprule
    &   & $\textbf{MJE}(\downarrow)$ & $\textbf{MBAE}(\downarrow)$ & $\textbf{PCK}(\uparrow)$ \\
    \midrule
    \textbf{3D cart.} (base) &  & $0.41^{\pm 0.08}$ & $36.93^{\pm 6.74}$ & $0.25^{\pm 0.11}$ \\
    \midrule
    w/ gloss cont. &  & $\bf 0.40^{\pm 0.08}$ & $ 36.02^{\pm 6.95}$ & $\bf 0.29^{\pm 0.12}$\\
    \midrule
    & $\lambda$ &  &  &  \\
    \cmidrule(lr){2-2}
    \multirow{4}{*}{\makecell[l]{w/ SBERT cont.\\\scriptsize batch $= 64$}}
    & $\it 0.0001$ & $0.40^{\pm 0.08}$ & $36.68^{\pm 6.97}$ & $0.28^{\pm 0.12}$ \\
    & $\it 0.0005$ & $\underline{0.40}^{\pm 0.08}$ & $36.57^{\pm 6.88}$ & $\underline{0.28}^{\pm 0.12}$ \\
    & $\it 0.001$ & $0.40^{\pm 0.08}$ & $36.76^{\pm 6.76}$ & $0.28^{\pm 0.13}$ \\
    & $\it 0.005$ & $0.41^{\pm 0.08}$ & $38.14^{\pm 6.67}$ & $0.26^{\pm 0.11}$ \\
    & $\it 0.01$  & $0.42^{\pm 0.08}$ & $38.93^{\pm 6.56}$ & $0.26^{\pm 0.11}$ \\
    & $\it 0.1$   & $0.44^{\pm 0.09}$ & $39.52^{\pm 7.14}$ & $0.24^{\pm 0.11}$ \\
    \midrule
    & batch &  &  &  \\
    \cmidrule(lr){2-2}
    \multirow{2}{*}{\makecell[l]{w/ SBERT cont.\\\scriptsize $\lambda = 0.001$}} 
    & \it{128} & $0.43^{\pm 0.08}$ & $39.06^{\pm 6.83}$ & $0.25^{\pm 0.11}$ \\
    & \it{256} & $0.42^{\pm 0.08}$ & $38.97^{\pm 7.04}$  & $0.25^{\pm 0.11}$ \\
    \bottomrule\toprule
    \textbf{quaternions} & & $0.44^{\pm 0.08}$ & $\underline{35.66}^{\pm 7.39}$ & $0.22^{\pm 0.10}$ \\
    \midrule
    w/ gloss cont. &  & $0.42^{\pm 0.09}$ & $\bf 34.69^{\pm 7.19}$ & $0.26^{\pm 0.12}$ \\
    \midrule
    & $\lambda$ &  &  &  \\
    \cmidrule(lr){2-2}
    \multirow{3}{*}{\makecell[l]{w/ SBERT cont.\\\scriptsize batch $= 64$}}
    & $\it 0.05$ & $0.44^{\pm 0.08}$ & $36.54^{\pm 7.98}$ & $0.24^{\pm 0.11}$ \\
    & $\it 0.1$ & $0.43^{\pm 0.09}$ & $36.36^{\pm 7.23}$ & $0.25^{\pm 0.12}$ \\
    & $\it 1$ & $0.43^{\pm 0.08}$ & $36.70^{\pm 7.08}$ & $0.25^{\pm 0.11}$ \\
    \bottomrule    
  \end{tabular}
\end{table}

\begin{figure}[h]
  \centering
  \includegraphics[width=\linewidth]{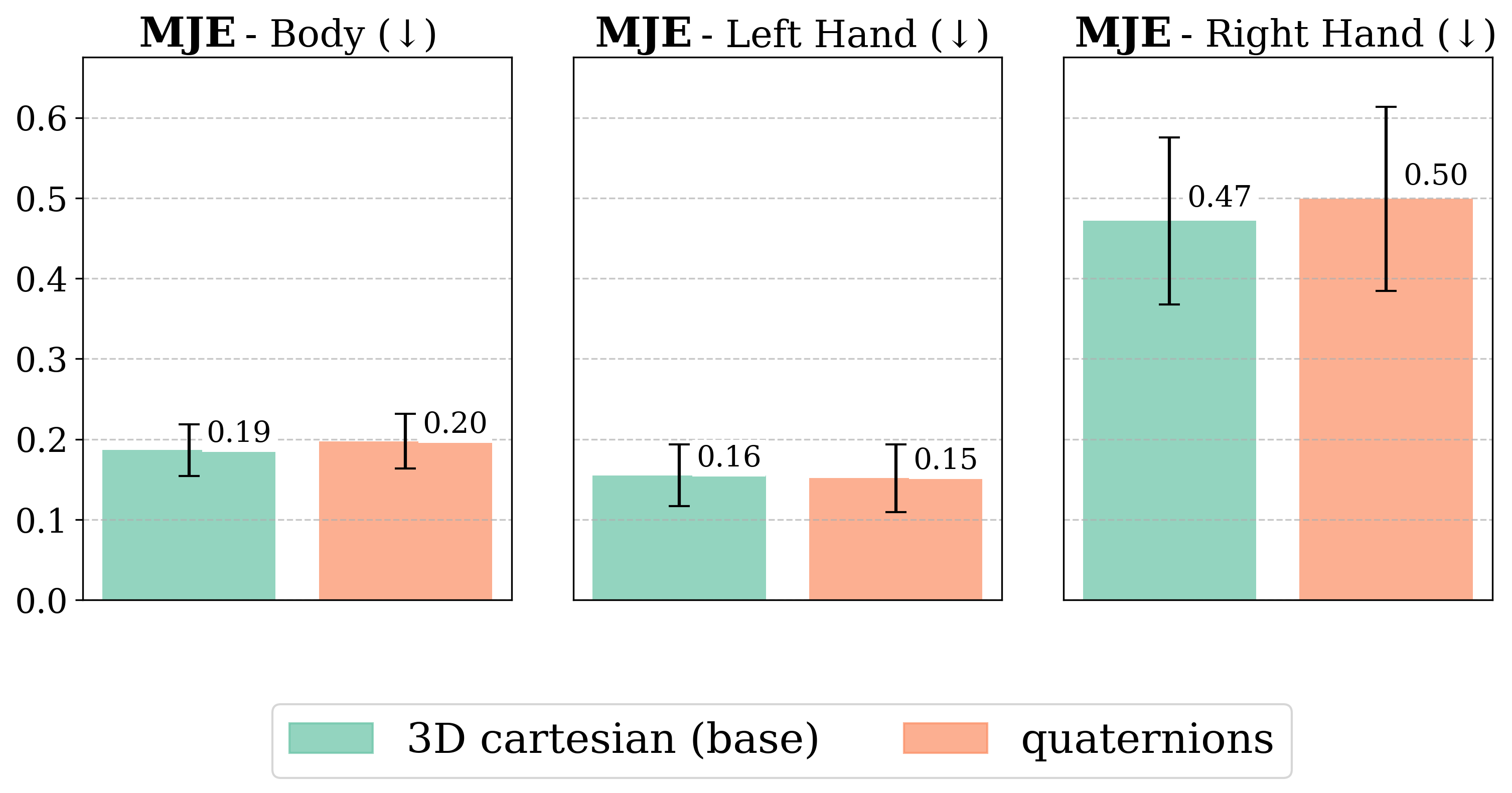}
  \vspace{-0.7cm}
  \caption{Bar plots of MJE per skeletal part on \textsc{Phoenix14T} test set between the base PT model and its quaternion-based variant.}
  \vspace{-0.4cm}
  \label{fig:base_vs_quat}
\end{figure}

As shown in Table \ref{tab:results}, encoding poses through bone rotations using quaternions---paired with geodesic loss optimization (see the "quaternions" row)---leads to a slight relative reduction ($-3\%$) in mean angular error compared to the baseline approach using joint 3D Cartesian positions and MSE loss ("3D cart (base)" row). 
This specific improvement aligns with the objectives of geodesic loss, which better respects the manifold structure of rotations. Qualitative analysis also reveals that the quaternion-based model often produces crisper and more distinct manual articulations, while the baseline tends to generate smoothed, averaged motions (see Figure \ref{fig:qualitative_gt_vs_base_vs_quat}). These results are consistent with findings from (\citet{signidd2025}), which emphasize the benefits of modeling bone orientations.\\
\indent However, using rotations instead of positions leads to diminished performance on standard joint-based metrics, such as  MJE, especially noticeable for the dominant right hand (Figure \ref{fig:base_vs_quat}). This performance drop is also reflected in lower PCK scores.

\subsubsection{PT Model with Contrastive Objectives}

Integrating a contrastive loss into the baseline Progressive transformers model consistently improves performance across all evaluated metrics. In particular, the use of $\mathcal{L}_{{\text{\scriptsize GlossSupCont}}}$ results in a 16\% relative improvement in PCK ("w/ gloss cont." in Table \ref{tab:results}), from 0.25 to 0.29, while the SBERT embeddings-based variant achieves a 12\% relative increase (“w/ SBERT cont.” row), from 0.25 to 0.28. These results are in line with previous studies such as (\citet{walsh2024}) and (\citet{signstokenszuo2024}) which reduce pose representation space density through vector quantization.\\
\indent Interestingly, the model trained with $\mathcal{L}{\text{\scriptsize SBERTSupCont}}$ performs slightly worse than the one using $\mathcal{L}{\text{\scriptsize GlossSupCont}}$. A likely explanation lies in the nature of the supervisory signals: SBERT-based supervision induces a smoother, more continuous embedding structure by aligning pose sequence similarities with sentence embedding similarities in $[0,1]$, whereas gloss-based supervision relies on binary similarity labels from shared glosses. This discretization may lead to stronger clustering effects, aiding training convergence and generalization.\\
\indent It would be beneficial to evaluate these approaches using metrics that better reflect semantic or linguistic intelligibility. Given that stylistic variation still exists in the evaluation data, standard positional metrics may not fully capture the comprehensibility of a sign. A generated sign may remain highly intelligible despite significant deviation from a reference pose, suggesting that current metrics might underestimate improvements that matter most for end-users.

\subsubsection{Combining Quaternions Pose Encoding and Contrastive Losses}

Combining quaternion-based pose encoding with contrastive training objectives compensates for the positional metric degradation observed when using geodesic loss alone. 
This combination also leads to further improvements in angular accuracy.  Specifically, training the quaternion-based model with $\mathcal{L}_{{\text{\scriptsize GlossSupCont}}}$ results in an additional 1° reduction in mean angular error,  a 4\% drop in MJE (from 0.44 to 0.42), and an 18\% increase in PCK (from 0.22 to 0.26) relative to the same model trained without contrastive objective.\\
\indent These findings suggest that integrating rotational encoding with angular-loss objectives and semantic-aware contrastive losses effectively addresses variability in signer morphology and style, ultimately producing clearer and more consistent sign motions.

\section{Conclusions}
We introduced and explored two complementary strategies to improve the classical Progressive Transformers model for sign language production, focusing on mitigating the impact of morphological and stylistic variability among signers. Our experiments, conducted on the \textsc{Phoenix14T} dataset, demonstrate that: (1) Encoding skeletal poses using bone rotations (quaternions) and optimizing them with a geodesic loss leads to more distinct angular motions, particularly for hand and finger articulations; (2) Augmenting the decoder with a contrastive loss that structures self-attention embeddings yields consistent improvements across all metrics, especially when using shared glosses to define positive sequence pairs; (3) Combining both methods results in further gains in angular precision while preserving joint position accuracy. These results advocate for the systematic inclusion of skeletal structure and rotation-aware representations, along with semantic-guided contrastive learning, in future SLP model training pipelines.

\indent As future work, we plan to incorporate back-translation metrics, such as BLEU, to more accurately evaluate the semantic intelligibility of generated sign sequences, beyond purely spatial or angular error measures. In addition, we aim to refine the contrastive supervision strategy by leveraging sentence Transformer embeddings to define positive and negative sequence pairs based on semantic similarity thresholds. This approach could offer a hybrid between the current discrete gloss-based method and the continuous SBERT-based formulation, potentially improving the alignment of the learned pose representations with semantic meaning.\\
Finally, given the known limitations of the \textsc{Phoenix14T} dataset, future work will involve evaluating our approach on the more recent \textsc{Mediapi-RGB} French Sign Language dataset, which offers greater diversity and scale, with over 86 hours of video \cite{mediapirgb2024}.

\begin{acks}
This work is part of \textit{Défi Inria COLaF}, which was financed by \textit{Plan National de Recherche en Intelligence Artificielle}.
\end{acks}

\bibliographystyle{ACM-Reference-Format}
\bibliography{main}

\end{document}